\documentclass{article}
\usepackage{ijcai20}

\usepackage{times}
\usepackage{soul}
\usepackage{url}
\usepackage[hidelinks]{hyperref}
\usepackage[utf8]{inputenc}
\usepackage[small]{caption}
\usepackage{graphicx}
\usepackage{amsmath}
\usepackage{amsthm}
\usepackage{booktabs}
\usepackage{tabularx}
\usepackage{comment}
\usepackage{amssymb}
\usepackage[boxed,ruled,commentsnumbered]{algorithm2e}

\usepackage{pifont}
\usepackage[caption=false]{subfig}
\newcommand{\citet}[1]{\citeauthor{#1} \shortcite{#1}}
\usepackage{mathtools}
\usepackage{multirow}

\makeatletter
\newcommand{\printfnsymbol}[1]{%
  \textsuperscript{\@fnsymbol{#1}}%
}
\makeatother

\title{Dynamic Knowledge Distillation for Black-box Hypothesis Transfer Learning}

\iftrue
\author{
Yiqin Yu$^1$\thanks{equal contribution}
\and
Xu Min$^1$\printfnsymbol{1}
\and
Shiwan Zhao$^1$\and
Jing Mei$^1$\and
Fei Wang$^2$\and
Dongsheng Li$^1$\and
Kenney Ng$^3$\And
Shaochun Li$^1$
\affiliations
$^1$IBM Research China, Beijing, China\\
$^2$Department of Healthcare Policy and Research, Cornell University, USA\\
$^3$IBM T.J. Watson Research Center, MA, USA\\
\emails
\{yuyiqin, minxux, zhaosw, meijing, ldsli, lishaoc\}@cn.ibm.com,
few2001@med.cornell.edu,
Kenney.Ng@us.ibm.com
}
\fi
\begin{document}

\maketitle

\begin{abstract}
In real world applications like healthcare, 
it is usually difficult to build a machine learning prediction model that works universally well across different institutions.
At the same time, the available model is often proprietary, i.e., neither the model parameter nor the data set used for model training is accessible. 
In consequence, leveraging the knowledge hidden in the available model (aka. the hypothesis) and adapting it to a local data set becomes extremely challenging. 
Motivated by this situation, in this paper we aim to address such a specific case within the hypothesis transfer learning framework, in which 1) the source hypothesis is a black-box model and 2) the source domain data is unavailable.
In particular, we introduce a novel algorithm called dynamic knowledge distillation for hypothesis transfer learning (dkdHTL). 
In this method, we use knowledge distillation with instance-wise weighting mechanism to adaptively transfer the ``dark'' knowledge from the source hypothesis to the target domain.
The weighting coefficients of the distillation loss and the standard loss are determined by the consistency between the predicted probability of the source hypothesis and the target ground-truth label.
Empirical results on both transfer learning benchmark datasets and a healthcare dataset demonstrate the effectiveness of our method.
\end{abstract}

\section{Introduction}

Along with the accumulation of digital data in healthcare, machine learning algorithms have been widely used to build numerous models to generate medical insights and improve healthcare practice for disease prevention, diagnosis, treatments and prognosis. 
However, while training machine learning models to account for complex biological phenomena and disease conditions, the limited availability of large amount of patient data becomes a bottleneck.
This leads to the difficulty of building a machine learning model, e.g., a risk prediction model, that is universally applicable for different patient cohorts and raises the requirement of leveraging knowledge from other cohorts.
For example, for predicting in-hospital mortality in acute coronary syndrome patients, several models have already been built based on different patient cohorts to capture their characteristics on target disease conditions \cite{granger2003predictors,li2018risk}. 
However, when building a risk model on a new cohort, the researcher is always required to evaluate whether knowledge from aforementioned models can be used.
On the other hand, the available model is often proprietary in real world applications, which means neither the model parameter nor the data used to train the model is accessible. 
This often happens when the data set and the model are highly valuable or legally restricted for distribution.
In this case, leveraging the knowledge hidden in available models and adapting it to a local data set becomes extremely challenging.

Transferring knowledge from a related \textit{source} domain to improve the learning performance in a \textit{target} domain is the major focus of Transfer Learning (TL) \cite{zhuang2019comprehensive}. 
Most of the TL works are devoted to do data-driven transfer, i.e., transfer by mitigating domain shifts based on instance distributions or feature distributions \cite{Fernandes2019}, which assume the source domain data is available. 
While considering the absence of source data, Hypothesis Transfer Learning (HTL) provides theoretical guarantees to improve the learning performance on target data by leveraging knowledge from auxiliary hypotheses.
Here the auxiliary hypotheses can be classifiers or regressors originating from other learning tasks which are built on different domains.
In this paper, we specify our interest on cross-domain knowledge transfer when 1) the target learning task can only leverage the source hypothesis as a black-box function and 2) the source domain data set is unavailable. 
We formalize our work under the theoretical framework of HTL and treat the source black-box hypothesis as the only knowledge from the source domain.

Besides theoretical works on cross-domain knowledge transfer, many practical knowledge transfer methods have been proposed with various intuitions. 
Unfortunately, the restricted access to internal parameters of the source model in our situation prevent us from transferring knowledge from intermediate layers such as in \cite{romero2014fitnets}.
This motivates us to use Knowledge Distillation (KD) \cite{hinton2015distilling}, an efficient method to distill ``dark'' knowledge from a teacher model by raising the temperature of its final softmax function, to solve the HTL problem.
Two issues need to be addressed before we equip KD into our framework. 
First, KD is originally used under circumstances of building smaller, faster models by ``compressing'' large, complex models.
KD only forces the student model to imitate the teacher model as much as possible, without dealing with any adaptation problem.
Second, in our situation the source hypothesis can only be visited as a black-box function, which means we can only get the predicted probability vector of the source model, and we cannot obtain the logits needed to calculate the soften probabilities for the source model.



Thereby in this paper, we address the above problems by introducing a novel algorithm called dynamic knowledge distillation for hypothesis transfer learning (dkdHTL). 
In this method, we use knowledge distillation with instance-wise weighting mechanism to adaptively transfer the ``dark'' knowledge of the source hypothesis to the target domain.
The KD method is customized to use predicted probabilities rather than the logits of source models.
The coefficient weights of the distillation loss and the standard loss are determined by the discrepancy between the predicted probability of the source hypothesis and the target ground-truth label. 
To the best of our knowledge, this is the first work that addresses the black-box HTL problem via instance-wise dynamic knowledge distillation.
The main contributions of this paper are highlighted as follows:

\begin{itemize}
\item We formalize our work under the theoretical framework of HTL and specialize it with a single black-box source hypothesis.
\item We propose the dynamic knowledge distillation with instance-wise weighting mechanism (dkdHTL) as a concrete algorithm for the HTL problem. 
\item Our method dkdHTL achieves promising empirical results on both transfer learning benchmark datasets 
and a healthcare dataset. 
\end{itemize}

\section{Related Work}
\subsection{Hypothesis Transfer Learning}
HTL provides generalized theoretical guarantees for finding optimal transfer parameters of source hypotheses. 
One main task of HTL is to figure out which hypotheses are helpful given a collection of source hypotheses \cite{wang2016learning,Kuzborskij2017}.
\citet{Kuzborskij2013} analyzes the stability of transferring a single hypothesis to the target domain based on least-squares with biased regularization. It assumes that the source hypothesis must be a linear predictor living in the same space of the target predictor, which prevents the use of the source hypothesis as a black-box function. 
Another theoretical work \cite{perrot2015theoretical} addresses HTL in the context of supervised regularized metric learning by using a biased regularization with source hypothesis.
Recently in \cite{Fernandes2019}, the HTL framework is generalized to include four learning tasks: regression, classification, learning to rank and recommender system. 
The authors use the structural similarity as transferable knowledge, which can be coefficients of a support vector machine for classification or the weak estimators and their associated importance for an AdaBoost model.
\citet{du2017hypothesis} proposes a practical HTL algorithm which predefines a transformation function and treat it as an input of the learning algorithm. 

\subsection{Knowledge Transfer}
Among general knowledge transfer works, KD \cite{hinton2015distilling} is an efficient approach to build smaller, faster models by ”distilling” knowledge from large, complex models. 
There have been studies using KD in the context of transfer learning and domain adaptation when the source domain data is not available.
For example, \citet{ao2017fast} utilizes a generalized distillation framework (an extension of KD) to leverage the knowledge from the source domain by using both labeled and unlabeled target data.
Besides, \citet{nayak2019zero} addresses the no training data problem in KD by synthesizing the crafted samples from the source model and using them as surrogates to train the target model. 
In spite of using KD, \citet{ahn2019variational} proposes a more general knowledge transfer framework through maximizing mutual information between two networks based on the variational information maximization technique.
Another method \cite{chidlovskii2016domain} extends feature corruptions and their marginalization to solve the domain adaptation problem. 
In healthcare applications, \citet{hong2018rdpd} addresses cross-domain knowledge transfer on healthcare data sets, by jointly optimizing the combined loss of attention imitation and target imitation during KD. 
\citet{mei2019knowledge} discusses three knowledge transfer approaches for developing risk prediction models on electronic health record data, including injecting knowledge to input features, to the objective functions and to output labels.

\section{Black-box Hypothesis Transfer Learning}
Transfer learning aims to improve the learning performance in the \textit{target} domain by transferring knowledge from the \textit{source} domain. 
A domain $\mathcal{D}$ usually contains an input space $\mathcal{X}$ and an output space $\mathcal{Y}$, practically observed by a number of instances in $\mathcal{X}$ and labels from $\mathcal{Y}$. 
In this paper, the $\textit{source}$ domain data is not visible.
We define the $\textit{target}$ domain as $\mathcal{D} = (\mathcal{X},\mathcal{Y})$ and corresponding training set as $D^t = \{(\mathbf{x}_i, \mathbf{y}_i) \mid i = 1, \cdots, n\}$ drawn i.i.d. from $\mathcal{X} \times \mathcal{Y}$. 
A hypothesis where the knowledge can be transferred from is a function $f \in \mathcal{F}$ that maps $\mathcal{X}$ to the set $\mathcal{Y}$, where $\mathcal{F} \subseteq \mathcal{Y}^{\mathcal{X}}$ is the hypothesis space. 
For the $\textit{source}$ domain, the model structure and parameters of the hypothesis $f^s \in \mathcal{F}$ are unknown. 
The only visible knowledge from $f^s$ is its predicted probability on the target data.
In general, we will transfer knowledge from $(\varnothing, f^s)$ to $(D^t, f^t)$, where $\varnothing$ means empty source data.

\subsection{Hypothesis Transfer Learning Framework}
With a non-negative loss function $\ell: \mathcal{Y} \times \mathcal{Y} \mapsto \mathbb{R}_+$, we denote by $\ell(f(\mathbf{x}),\mathbf{y})$ the loss of hypothesis $f$ on instance $(\mathbf{x}, \mathbf{y})$. 
The risk of $f^t$, with respect to the target probability distribution $\mathcal{D}$, and the empirical risk measured on $D^t$ are defined as
\begin{equation}
\begin{aligned}
\label{eqn:risks}
    R_{\mathcal{D}}(h)\vcentcolon=\mathbb{E}_{(\mathbf{x}, \mathbf{y}) \sim \mathcal{D}}[\ell(f^t(\mathbf{x}), \mathbf{y})], \\
    \hat{R}_{D^t}(h)\vcentcolon=\frac{1}{n} \sum_{i=1}^{n} \ell\left(f^t\left(\mathbf{x}_{i}\right), \mathbf{y}_{i}\right).
\end{aligned}
\end{equation}
In order to inject knowledge from $f^s$ while minimizing the above risks for $f^t$, 
we formalize the HTL problem within the framework of Regularized Empirical Risk Minimization (ERM):
\begin{equation}
    \label{eqn:target_task}
\hat{f^t}= \underset{f^t}{\arg \min }\frac{1}{n}  \sum_{i=1}^{n} \alpha \ell(f^t(\mathbf{x}_{i}), \mathbf{y}_{i}) + \beta \Omega (f^t),
\end{equation}
where $\Omega(f^t)$ is the regularization term on the target model $f^t$ to penalize its inconsistency with the knowledge contained in the source model $f^s$, and $\alpha, \beta$ are weighting coefficients deciding what proportion of knowledge from $f^s$ to be transferred.


In general, the regularization term for ERM can be in various forms.
For example, if the source model $f^s$ is a white-box function and is in the same hypothesis space as the target model, it can be defined as the distance between two functions
$\Omega(f^t) = \|f^t-f^s\|$ \cite{Kuzborskij2013}, where $\|\cdot\|$ is a norm function defined in space $\mathcal{F}$.
In the case of multiple auxiliary hypotheses, hypothesis-wise weighting should be considered in this regularization term \cite{Kuzborskij2017}, e.g., $\Omega(f^t) = \|f^t - f^s_{\gamma}\|$, where $f^s_{\gamma} =\sum_i \gamma_i f^s_i $ is an optimal combination of source hypotheses.
Besides, given the source domain data,
\citet{wang2019transfer} define regularization as the performance gap between the source domain and the target domain $\Omega(f^t) =(\mathcal{L}_{D^{s}}^{{\Gamma}^s}({f}^t) - \mathcal{L}_{D^{s}}^{{\Gamma}^s}(f^s))+(\mathcal{L}_{D^{t}}^{{\Gamma}^t}(f^s) - \mathcal{L}_{D^{t}}^{{\Gamma}^t}({f}^t))$ where ${\Gamma}^s$ and ${\Gamma}^t$ are instance-wise weighting parameters for data in the source domain and the target domain, respectively. 
Different from the above situations, we have no access to either the parameters of the source hypothesis or the source domain data.
To overcome this challenge, we will resort to knowledge distillation and define $\Omega(f^t)$ using the distillation loss.


\subsection{Dynamic Knowledge Distillation with Instance-wise Weighting Mechanism}
In this section we propose dynamic knowledge distillation as a concrete algorithm of black-box hypothesis transfer learning formulated in Eq. \ref{eqn:target_task}. 
For the sake of simplicity, we omit the index of $(\mathbf{x}_i, \mathbf{y}_i)$ and directly use $(\mathbf{x}, \mathbf{y})$ to denote the $i$-the sample. 
We also use $\mathbf{p}^t$ and $\mathbf{p}^s$ to denote the output probability vector of $f^t(\mathbf{x})$ and $f^s(\mathbf{x})$, respectively.
We can formulate our regularized loss function as
\begin{equation}
\label{eq:loss_kd}
    \mathcal{L}(\mathbf{x}, \mathbf{y}; \mathbf{w}) = \alpha  \mathcal{L}_1(\mathbf{p}^t, \mathbf{y}) + \beta  \mathcal{L}_2(\mathbf{p}^t, \mathbf{p}^s; T).
\end{equation}

The first term is the standard loss for the supervised learning task corresponding to the first part in Eq. \ref{eqn:target_task}; The cross-entropy loss is often used for a classification problem. 
\begin{equation}
\label{eq:loss_1}
    \mathcal{L}_1(\mathbf{p}^t, \mathbf{y}) = H(\mathbf{y}, \mathbf{p}^t) \vcentcolon = -\sum_{j=1}^{C} y_j \log p^t_j,
\end{equation}
where $H(\cdot, \cdot)$ is the cross-entropy function, $C$ is the number of classes, $y_j$ indicates whether the sample belongs to class $j$, and $p^t_j$ is the probability of the sample belonging to class $j$ which is computed by the target model.

Inspired by knowledge distillation \cite{hinton2015distilling}, the second term is the distillation loss which aims to leverage knowledge from $f^s$. 
The source model can generate a large probability for the true class while small probabilities for other classes.
The dark knowledge underlying the small probabilities can be distilled using a softmax with a high temperature $T>1$:
\begin{equation}
\label{eq:softmax_t}
    \tilde{p}_j = \sigma(z_j; T) \vcentcolon = \frac{\exp(z_j/T)}{\sum_j \exp(z_j/T) }.
\end{equation}
However in the case of a black-box hypothesis $f^s$, we cannot directly get the logits $z_j$ needed to compute the soften probability $\tilde{p}_j$.
We tackle this issue by first reverting the probability back to the logits $z_j = \log (p_j) + c$ ($c$ is a constant number) and then plug $z_j$ into Eq. \ref{eq:softmax_t}.
As a result, specially for the black-box source hypothesis, we can soften the original predicted probability $p_j$ using
\begin{equation}
\label{eq:softmax_t_black}
    \tilde{p}_j = \frac{\exp(\log (p_j)/T)}{\sum_j \exp(\log (p_j)/T) },
\end{equation}
where the intermediate constant $c$ is cancelled out. 
Thus, we can calculate the distillation loss $\mathcal{L}_2$ for regularization by  
\begin{equation}
\label{eq:loss_2}
    \mathcal{L}_2(\mathbf{p}^t, \mathbf{p}^s; T) = H(\tilde{\mathbf{p}}^s, \tilde{\mathbf{p}}^t) = -\sum_{j=1}^{C} \tilde{p}^s_j \log \tilde{p}^t_j,
\end{equation}
where  $\tilde{p}^t_j$ is computed by Eq. \ref{eq:softmax_t} for the target model $f^t$, while $\tilde{p}^s_j$ is computed by Eq. \ref{eq:softmax_t_black} for the black-box source model $f^s$.

\begin{figure}[t]
    \centering
    \includegraphics[width=0.48\textwidth]{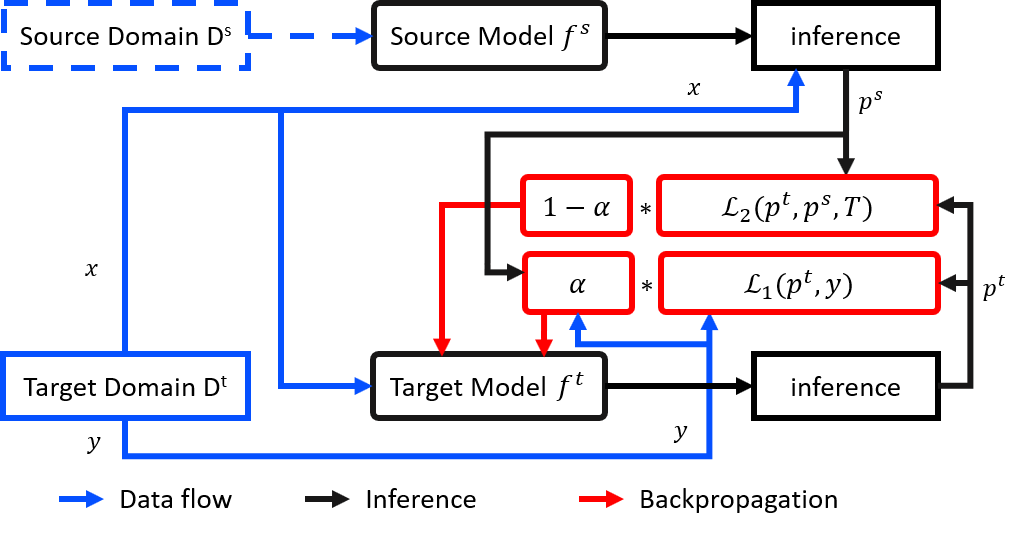}
    \caption{The framework of dynamic knowledge distillation for HTL. The back-propagation process of the target model $f^t$ shown with red dotted lines is produced by dynamic knowledge distillation, which considers losses from both of the hard labels $\mathcal{L}_1(\mathbf{p}^t, y)$ and the soft labels $\mathcal{L}_2(\mathbf{p}^t, \mathbf{p}^s; T)$ with instance-wise weighting mechanism controlled by $\alpha$.}
    \label{fig:arch}
\end{figure}

The most important contribution lies in that we propose an instance-wise weighting mechanism for the coefficients $\alpha$ and $\beta$ to avoid negative transfer.
We adjust the weights between the standard loss $\mathcal{L}_1$ and the distillation loss $\mathcal{L}_2$ based on a consistency score $S$ between the ground-truth label $\mathbf{y}$ and the probability $\mathbf{p}^s$ predicted by the source hypothesis $f^s$.
The consistency score $S$ is expected to be in the range $[0,1]$, and to be larger when $\mathbf{p}^s$ is closer to $\mathbf{y}$. 
A practical definition of such $S$ can be the exponential of the negative cross-entropy loss of $\mathbf{p}^s$ against $\mathbf{y}$:
\begin{equation}
\label{eq:consitence_score}
    S(\mathbf{y}, \mathbf{p}^s) = \exp(-H(\mathbf{y}, \mathbf{p}^s)) = \exp(\sum_{j=1}^{C}y_j \log(p^s_j) ).
\end{equation}
Finally, the instance-wise dynamic weighting mechanism is defined as
\begin{equation}
    \begin{aligned}
	\label{eq:weights}
		\alpha & = \lambda + \delta (1 - S(\mathbf{y}, \mathbf{p}^s)),\\
		\beta & = 1 - \alpha,
    \end{aligned}
\end{equation}
where $\lambda$ and $\delta$ are two hyperparameters satisfying $0\leq \lambda \leq 1$, $0\leq \delta \leq 1$, and $0\leq \lambda + \delta \leq 1$.
The coefficient $\alpha$ will be instance-wise adjusted in the range $[\lambda, \lambda + \delta]$.
Through this mechanism, the target learning task will focus more on the standard loss when $f^s$ generates inconsistent predictions with the ground-truth label, and hence avoid the negative transfer problem. 
The overview of our dynamic knowledge distillation is illustrated in Figure \ref{fig:arch}.
We also summarize the detailed training procedure of $f^t$ in Algorithm \ref{alg:dkdhtl}.

\begin{algorithm}
    \caption{Dynamic KD for HTL (dkdHTL)}
    \label{alg:dkdhtl}
    \SetAlgoNoLine
    \SetKwInOut{Input}{\textbf{Input}}\SetKwInOut{Output}{\textbf{Output}}
    \Input{
        The source hypothesis $f^s$, target domain data $D^t$ and hyperparameters for dynamic KD: 
         $\lambda$,
         $\delta$,
         $T$.
        }
    
    \textbf{Initialize} target model parameters $\mathbf{w}$ randomly;
    
    \For {$i \in MaxInteration$}{
        Take a batch of data $\mathcal{B}_i$ from $D^t$;
        
        \For{each instance $(\mathbf{x}, \mathbf{y})\in \mathcal{B}_i$}{
            Compute predicted probability ${\mathbf{p}}^t$ and ${\mathbf{p}}^s$;
            
            Compute soften probability $\tilde{\mathbf{p}}^t$ and $\tilde{\mathbf{p}}^s$ by Eq. \ref{eq:softmax_t} and \ref{eq:softmax_t_black} with temperature $T$;
            
            Compute consistence score $S(\mathbf{y}, \mathbf{p}^s)$ by Eq. \ref{eq:consitence_score};
        
            Compute the standard loss $\mathcal{L}_1(\mathbf{p}^t, \mathbf{y})$ and the distillation loss $\mathcal{L}_2(\mathbf{p}^t, \mathbf{p}^s)$ by Eqs. \ref{eq:loss_1} and \ref{eq:loss_2};
        
            Compute weighting coefficients $\alpha$ and $\beta$ by Eq. \ref{eq:weights} with hyperparameters $\lambda$ and $\delta$;
        
            Compute the total loss $\mathcal{L}$ by Eq. \ref{eq:loss_kd};
        }
        
        Compute gradient $\nabla\mathcal{\bar{L}}$ against $\mathbf{w}$ on the averaged total loss $\mathcal{\bar{L}}$  in batch $\mathcal{B}_i$;
        
        Update $\mathbf{w}$ using an Adam optimizer;
    }
    
    \Output{
        The model parameters $\mathbf{w}$ for $f^t$.
    }
\end{algorithm}

\section{Experiments}

We evaluate our algorithm on two popular transfer learning benchmark datasets including MNIST \cite{lecun1998gradient} and Office-31 \cite{saenko2010}, and one healthcare dataset MIMIC-III \cite{johnson2016mimic} which is used in a diverse range of healthcare related learning tasks. 
As we don't find most recent work that have exactly the same setup with us, we will mainly use the following baselines as benchmark experiments. 

As formulated in Eq. \ref{eqn:target_task}, there are two main parts contributing to the performance: the knowledge from the source hypothesis, and the information from the target data. 
Accordingly, two baselines methods are designed:

\begin{itemize}
\item Source Hypothesis only (\textbf{SH}). Directly testing the black-box source hypothesis $f^s$ on the target data set.
\item Target Data only (\textbf{TD}). Directly training on the target training data set $D^t$ without leveraging the source hypothesis $f^s$.
\end{itemize}

Combining knowledge available from both $f^s$ and $D^t$, we have another two methods:
\begin{itemize}
\item Static KD for HTL (\textbf{skdHTL}). Implementation of a standard KD algorithm \cite{hinton2015distilling}, where the distillation coefficient is a fixed value, i.e, $\delta=0$.
\item Dynamic KD for HTL (\textbf{dkdHTL}). Implementation of our proposed instance-wise dynamic Knowledge distillation in Eq. \ref{eq:loss_kd}.
\end{itemize}

Performances on MNIST and Office-31 are evaluated from the perspective of classification accuracy. 
Performance on MIMIC-III are evaluated using several metrics due to the complexity of the learning requirements. 
Experiments on MNIST and MIMIC-III are implemented in TensorFlow, and experiments on Office-31 are implemented in PyTorch. 
All experiments are run with NVIDIA Tesla P100 GPUs. 

\subsection{Experiments on MNIST}
To validate that our algorithm can overcome domain shift, we specially create a modified version of MNIST which consists of a source and a target domain. 
Specifically, we can omit all samples of certain digits to curate the source domain.
For example, when omitting digit 3, we obtain a source domain with 49362, 4507 and 8990 samples in training, validation and test set, respectively.
Meanwhile, to curate the target domain, we randomly sample 10\% of all examples of MNIST, which gives us a target domain with 5500, 500 and 1000 samples in training, validation and test set, respectively.
We call this modified MNIST dataset as ``m-MNIST". 
There are three key properties of m-MNIST: 1) there are overlaps between source domain and target domain, 2) target domain has much fewer samples than source domain, 3) source domain sees fewer classes than target domain.
Therefore, m-MNIST can be regarded as a simulation of domain \textit{data shift}, more specifically the \textit{prior shift}, in domain adaptation. 

We first train a CNN model with 2 convolution layers on the source domain of m-MNIST removing digits 3, to serve as the source model $f^s$ to be transferred.
This model achieves an accuracy of 98.59\% on the source test set. 
However, as can be seen in Table \ref{tab:performance_m-MNIST}, it only obtains an accuracy of only 88.82\% on the target test set due to its misclassification on digits 3.
In fact, we can prove the source model is still very valuable since it provides a wealth of knowledge regarding to the other nine digits.
We choose a one-layer MLP model as the target model $f^t$, and the TD model reaches an accuracy of 86.50\%. 
If we applies skdHTL with $\lambda=0.5, T=2$, we get a negative transfer with a worse accuracy of 82.68\%. 
Nevertheless, we can improve the accuracy to 88.62\% using dkdHTL with $\lambda=0.1, \delta=0.9, T=2$.

Furthermore, if we omit more digits in the source domain, e.g., omitting both digits 3 and 7, the accuracy of the SH model will be further decreased to 77.44\%.
Given this imperfect source model, skdHTL leads us to a worse transfer result with an accuracy of 75.83\% .
Nevertheless, using dkdHTL we can still achieve a target model of accuracy as high as 88.42\%.
Besides, we also explore the setting where a two-layer CNN model is used as the target model $f^t$.
The TD model and skdHTL model give us a target model of accuracy 94.06\% and 79.15\% respectively, which are better than the setting with an MLP as $f^t$ as expected.
In the meantime, dkdHTL achieves the best accuracy of 94.25\%.

Following the third setting in Table \ref{tab:performance_m-MNIST}, we also investigate the effects of hyperparameters $\lambda$, $\delta$, and $T$ in dkdHTL.
According to Table \ref{tab:hyper_param}, we find that dkdHTL achieves the best accuracy of 96.68\% with $\lambda=\delta=0.5, T=3$.
More generally, a larger $\lambda$, which makes the target model focus more on the target data $D^t$, gives a better accuracy.
Meanwhile, a larger $\delta$, which lets the target model learn more from the source model $f^s$ when it is consistent with ground-truth, gives a better accuracy as well.
The temperature $T$ shows relatively mild effects on the accuracy.

\begin{table}[t!]
\caption{HTL experiments on m-MNIST. 
We show the accuracy (\%) on the test, validation and training set in the target domain for different transfer settings and methods.
skdHTL uses $\lambda=0.5, \delta=0, T=2$, while dkdHTL uses $\lambda=0.1, \delta=0.9, T=2$. 
The best test accuracy among TD, skdHTL and dkdHTL is shown in \textbf{bold}.
}
\label{tab:performance_m-MNIST}
\small
\centering
 \begin{tabular}{c|c|cccc} 
 \toprule
    SETTING & METHOD & test acc & val acc & train acc\\ 
    \midrule
    $f^s$: CNN w/o 3   & SH & {88.82} & - & - \\
    & TD & 86.50 & 87.38 & 91.31 \\
    $f^t$: MLP    & skdHTL & 82.68	& 82.45 & 86.02\\
    & dkdHTL & \textbf{88.62} & 87.77 & 92.06 \\
    \midrule
    $f^s$: CNN w/o 3,7 & SH & 77.44 & - & -\\
    & TD &  86.50 & 87.38 & 91.31 \\
    $f^t$: MLP & skdHTL & 75.83 & 73.57 & 79.13 \\
    & dkdHTL & \textbf{88.42} & 88.36 & 91.90 \\
    \midrule
    $f^s$: CNN w/o 3,7  & SH & 77.44 & - & -\\
    & TD & 94.06 & 95.07 & 98.47 \\
    $f^t$: CNN & skdHTL & 79.15 & 81.26 & 82.65 \\
    & dkdHTL & \textbf{94.26} & 92.90 & 97.65\\
 \bottomrule
 \end{tabular}
\end{table}

\begin{table}[t!]
    \caption{Test accuracy (\%) of dkdHTL with different hyperparameters $\lambda, \delta, T$.
    The best result is shown in \textbf{bold}.}
    \label{tab:hyper_param}
    \small
    \centering
    \begin{tabular}{cccccc}
    \toprule
    \multirow{2}*{$\lambda$} & \multirow{2}*{$\delta$} & \multicolumn{3}{c}{$T$} & \multirow{2}*{AVERAGE} \\
    \cline{3-5}
    ~ & ~ & 2. & 3. & 4. & ~ \\
    \midrule
    0.1 & 0.3 & 79.36 & 78.11 & 78.32 & 78.60 \\
0.1 & 0.5 & 80.50 & 81.02 & 80.29 & 80.60 \\
0.1 & 0.7 & 89.11 & 83.19 & 87.45 & 86.58 \\
\midrule
0.3 & 0.3 & 79.88 & 80.91 & 79.36 & 80.05 \\
0.3 & 0.5 & 85.58 & 84.23 & 82.68 & 84.16 \\
0.3 & 0.7 & 95.44 & 94.40 & 95.12 & 94.99 \\
\midrule
0.5 & 0.3 & 83.40 & 85.27 & 84.54 & 84.41 \\
0.5 & 0.5 & 95.12 & \textbf{96.68} & 96.27 & 96.02 \\
0.7 & 0.3 & 95.85 & 96.06 & 96.58 & {96.16} \\
    \midrule
    \multicolumn{2}{c}{AVERAGE} & {87.14} & 86.65 & 86.73 & -\\
    \bottomrule
    \end{tabular}
\end{table}

\subsection{Experiments on Office-31}
Office-31 \cite{saenko2010} is a standard dataset for visual transfer learning. 
It has three domains: Amazon (\textbf{A}), Webcam (\textbf{W}), and DSLR (\textbf{D}). 
Each domain contains 31 unbalanced classes with a total of 4,110 images.
Choosing different pairs of the \textit{source} domain and the \textit{target} domain, we construct six HTL tasks: A$\to$W, D$\to$W, W$\to$D, A$\to$D, D$\to$A, and W$\to$A. 
The character on the left side of $\to$ represents the source domain from  which the hypothesis $f^s$ is generated, while the right side means the target domain. 


Following state-of-the-art transfer learning/domain adaptation works such as \cite{zhang2019bridging}, we use \textbf{ResNet-50} \cite{he2016deep} with parameters fine-tuned from the model pre-trained on ImageNet \cite{russakovsky2015imagenet} for both $f^s$ and $f^t$. 
$f^s$ is first trained on the corresponding \textit{source} domain in advance, and then used as a black-box function during the training of $f^t$.  

\begin{table*}[t!]
    \caption{Classification accuracy (\%) on Office-31 with training:validation:test=8:1:1.}
    \label{tab:performance_office31_1}
    \small
    \setlength{\tabcolsep}{6.5mm}
    \begin{tabularx}{\textwidth}{c|ccccccc}
    \toprule
      METHOD & A$\to$W & D$\to$W & W$\to$D & A$\to$D & D$\to$A & W$\to$A & AVERAGE \\
    \midrule
    SH & 63.75 & 82.50 & 90.00 & 68.00 & 52.84 & 62.06 & 69.86 \\
    TD & \textbf{100.00} & \textbf{100.00} & \textbf{100.00} & {98.00} & 86.88 & 86.52 & 95.23  \\
    skdHTL & 100.00 & 98.75 & 100.00 & 98.00 & 86.88 & 85.82 & 94.91\\
    \textbf{dkdHTL1} & \textbf{100.00} & \textbf{100.00} & \textbf{100.00} & 98.00 & 87.23 & 86.88 & 95.35\\
    \textbf{dkdHTL2} & \textbf{100.00} & \textbf{100.00} & \textbf{100.00} & \textbf{100.00} & \textbf{100.00} & \textbf{87.94} & \textbf{96.28} \\
    \bottomrule
    \end{tabularx}
\end{table*}

\begin{table*}[t!]
    \caption{Classification accuracy (\%) on Office-31 with training:validation:test=5:1:4.}
    \label{tab:performance_office31_2}
    \small
    \setlength{\tabcolsep}{6.5mm}
    \begin{tabularx}{\textwidth}{c|ccccccc}
    \toprule
      METHOD & A$\to$W & D$\to$W & W$\to$D & A$\to$D & D$\to$A & W$\to$A & AVERAGE \\
    \midrule
    SH & 62.07 & 88.40 & 96.50 & 71.50 & 56.65 & 59.31 & 72.41 \\
    TD & \textbf{95.30} & 96.55 & 96.00 & \textbf{97.50} & 84.31 & 83.95 & \textbf{92.27}  \\
    skdHTL & 94.04 & 96.87 & 95.50 & 95.00 & 83.42 & 82.45 & 91.21\\
    \textbf{dkdHTL1} & 94.04 & \textbf{97.18} & 96.00 & 96.00 & 84.22 & 83.16 & 91.77\\
    \textbf{dkdHTL2} & 94.98 & 96.55 & \textbf{99.00} & 93.50 & \textbf{84.49} & \textbf{84.04} & 92.09\\
    \bottomrule
    \end{tabularx}
\end{table*}

We run the experiments on different partitions of Office-31 to evaluate the performance when the training data is large enough as well as when the training data is not sufficient. 
The first training-validation-test partition is 8:1:1 for each domain, and the second partition is 5:1:4, which has much fewer training instances than the first one.
Classification accuracy for different tasks with a large training set as well as a small training set are reported in Table \ref{tab:performance_office31_1} and Table \ref{tab:performance_office31_2}, respectively. 
Note that the performances of TD on all tasks decrease on small training set comparing to the large one, whereas the performances of SH don't strictly align with the size of the test set. 
For skdHTL models, hyper-parameters are set as $\lambda =0.5$, $\delta=0$ and $T=2$, following the common KD practice.
Two types of dkdHTL models with different hyperparameter settings are evaluated: 1) dkdHTL1 with $\lambda =0.5$, $\delta =0.5$ and $T=2$, and 2) dkdHTL2 with $\lambda =0.7$, $\delta=0.3$ and $T=4$. 

From Table \ref{tab:performance_office31_1}, we can observe that the two dkdHTL models consistently outperform almost all baselines.
Comparing with skdHTL, the dynamic knowledge distillation is more suitable in transfer learning tasks. 
With a smaller training partition in the target domain, the overall performances become degraded due to the insufficiency of training samples, as shown in Table \ref{tab:performance_office31_2}. 
For three adaptation tasks $W\to D$, $D \to A$, $W \to A$, dkdHTL2 still achieves the optimal performance (99.00\%, 84.49\%, 84.04\%), whereas it is inferior to the TD model in tasks $A \to W$ and $A \to D$. 
It is reasonable since different tasks may have different optimal hyperparameter settings, due to the heterogeneity of transfer tasks. 
It is worth noting that the task $D \to A$ is the most difficult task out of all six tasks, and dkdHTL2 always gets the highest classification accuracy in this task.
Overall, domain adaptation experiments on Office-31 validate the effectiveness of dkdHTL.



\subsection{Experiments on MIMIC-III benchmarks}
Based on MIMIC-III \cite{johnson2016mimic}, a freely accessible critical care database, a recent work \cite{harutyunyan2019multitask} constructed benchmark machine learning datasets.
In particular, an ``In-Hospital Mortality" (IHM) task was defined to predict whether an ICU patient will die at discharge given the first 48 hours observation of the ICU stay.

\begin{figure}[t]
    \centering
    \includegraphics[width=\linewidth]{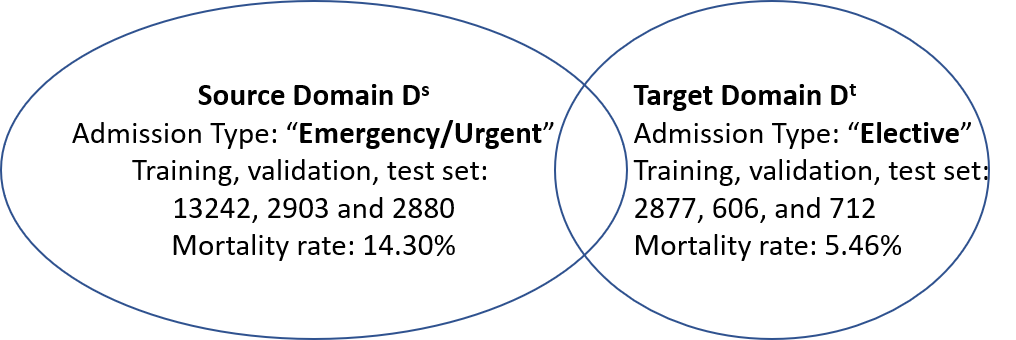}
    \caption{MIMIC-III IHM datasets are split into a source domain and a target domain according to the type of ICU admission.}
    \label{fig:mimic3_split}
\end{figure}

In our experiment, we first split the MIMIC-III data into two domains according to the type of ICU admission. 
As can be seen in Figure \ref{fig:mimic3_split}, the source domain contains patients with admission type ``Emergency/Urgent", while the target domain contains patients with admission type ``Elective".
The source domain has 13242, 2903 and 2880 samples in the training, validation and test set, respectively. 
Meanwhile the target domain has 2877, 606, and 712 samples in the training, validation and test set, respectively. 
Each sample has 48 timestamps of 76 features, with a label indicating mortality at discharge.

Note that there exists significant class imbalance in this IHM task.
In detail, the mortality rate in the source domain is 14.30\% and that in the target domain is only 5.46\% (this also reflects the domain difference).
Therefore with regard to model evaluation, we consider three measurements including accuracy, auROC (area under the Receiver Operating Characteristic curve) and auPRC (area under the Precision Recall Curve). 
To keep consistence with \citet{harutyunyan2019multitask}, we adopt bidirectional RNN with LSTM (Long-Short Term Memory) units to implement all networks.
The source model and target model have the same RNN architecture, both with 16 hidden units in the bidirectional LSTM layer.

We train a source model on the source domain, which generates accuracy of 88.68\%, auROC of 84.52\% and auPRC of 48.72\%, on its test set. 
Note that this result is very close to the performance reported by \citet{harutyunyan2019multitask} which is trained on the whole dataset.
According to the results reported in Table \ref{tab:performance_mimic3}, the source model is a good teacher model on the target domain, with accuracy of 94.24\%, auROC of 87.26\% and auPRC 36.39\%.
In contrast, the TD model achieves the worst performance considering auROC of 84.06\% and auPRC of 29.31\%. 
The skdHTL model with $\lambda=0.5, \delta=0, T=2$ improves the performance of TD model to auROC of 86.46\% and auPRC of 33.10\%. 
The dkdHTL model with $\lambda=0.1, \delta=0.1, T=4$ brings us a better target model with auROC of 86.70\% and auPRC of 34.08\%, and with smaller standard deviations.
This hyperparameter setting for dkdHTL is chosen from $\lambda,\delta \in [0.1,0.3,0.5,0.7,0.9], T \in [2,3,4]$  by grid search.
We find that the dkdHTL model is slightly inferior to the SH model in terms of auROC and auPRC. 
It is because there is inherently a gap between the student model and the black-box teacher model in knowledge distillation.
In fact, if the detailed parameters $\mathbf{w}_s$ of $f^s$ is available, stronger regularization terms such as $\|\mathbf{w} - \mathbf{w}_s\|$, can be added to our loss function in order to force the target model to mimic the source model more closely.

\begin{table}[t!]
    \caption{Real-world medical experiments of the In-hospital mortality prediction task on MIMIC-III benchmarks. 
    All experiments are run for five times, and the average accuracy, auROC and auPRC (\%) on the test set in the target domain are reported with standard deviation in the following bracket. 
    The best result among TD, skdHTL and dkdHTL is shown in \textbf{bold}.}
    \label{tab:performance_mimic3}
    \small
    \centering
    \begin{tabular}{c|ccc}
    \toprule
    METHOD & accuracy & auROC & auPRC \\
    \midrule
    SH &  94.24 (0.00) & 87.26 (0.00) & 36.39 (0.00) \\
    \midrule
    TD & 94.52 (0.34) & 84.06 (1.03) & 29.31 (2.50) \\
    skdHTL & 94.78 (0.44) & 86.46 (0.60) & 33.10 (4.37) \\
    dkdHTL & \textbf{94.94 (0.22)} & \textbf{86.70 (0.57)} & \textbf{34.08 (1.60)} \\
    \bottomrule
    \end{tabular}
\end{table}

\section{Conclusion}
In this paper, we address a HTL problem whereas the target learning task can leverage the source hypothesis only as a black-box function and source domain data is completely unavailable. 
An instance-wise dynamic weighting mechanism is incorporated into the knowledge distillation method to form a concrete practical algorithm for this specific HTL problem. 
Extensive empirical experiments on both transfer learning benchmark datasets and a healthcare dataset prove the effectiveness of our approach.

\bibliographystyle{named}
\bibliography{reference}

\end{document}